 \documentclass[11pt,bezier,]{article}

\newtheorem{proposition}{Proposition}

\usepackage{amsmath}
\usepackage{amssymb}
\usepackage{amsfonts}
\usepackage{mathrsfs}
\usepackage{subfigure}
\usepackage{epsfig}
\usepackage{graphicx}
\usepackage{color}

\usepackage[linesnumbered,ruled,]{algorithm2e}

\newcommand{\myBox}{\squareforqed}

\newenvironment{proof}{\par\noindent{\bf Proof:}}{\hfill$\myBox$\par}

\newcommand{\blackhole}[1]{}

\newfont{\bbfv}{   msbm5}                      
\newfont{\bbfvi}{  msbm6}                      
\newfont{\bbfvii}{ msbm7}                      
\newfont{\bbfviii}{msbm8}                      
\newfont{\bbfix}{  msbm9}                      
\newfont{\bbfx}{   msbm10}                     
\newfont{\bbfxi}{  msbm10 scaled\magstephalf}  
\newfont{\bbfxii}{ msbm10 scaled\magstep1}     
\newfont{\bbfxiv}{ msbm10 scaled\magstep2}     
\newfont{\bbfxvii}{msbm10 scaled\magstep3}     
\newfont{\bbfxx}{  msbm10 scaled\magstep4}     
\newfont{\bbfxxv}{ msbm10 scaled\magstep5}     

\def\bbf#1{{\relax\rm
\ifdim\the\fontdimen6\the\font<7pt         
 \mbox{\bbfv #1}%
\else\ifdim\the\fontdimen6\the\font<7.6pt  
 \mbox{\bbfvi #1}%
\else\ifdim\the\fontdimen6\the\font<8.25pt 
 {\ifmmode\mathchoice{\mbox{\bbfvii #1}}
  {\mbox{\bbfvii #1}}{\mbox{\bbfvi #1}}
  {\mbox{\bbfv #1}}\else{\bbfvii #1}\fi}%
\else\ifdim\the\fontdimen6\the\font<8.85pt 
 {\ifmmode\mathchoice{\mbox{\bbfviii #1}}
  {\mbox{\bbfviii #1}}{\mbox{\bbfvi #1}}
  {\mbox{\bbfv #1}}\else{\bbfviii #1}\fi}%
\else\ifdim\the\fontdimen6\the\font<9.7pt  
 {\ifmmode\mathchoice{\mbox{\bbfix #1}}
  {\mbox{\bbfix #1}}{\mbox{\bbfvi #1}}
  {\mbox{\bbfv #1}}\else{\bbfix #1}\fi}%
\else\ifdim\the\fontdimen6\the\font<10.5pt 
 {\ifmmode\mathchoice{\mbox{\bbfx #1}}
  {\mbox{\bbfx #1}}{\mbox{\bbfvii #1}}
  {\mbox{\bbfv #1}}\else{\bbfx #1}\fi}%
\else\ifdim\the\fontdimen6\the\font<11.4pt 
 {\ifmmode\mathchoice{\mbox{\bbfxi #1}}
  {\mbox{\bbfxi #1}}{\mbox{\bbfviii #1}}
  {\mbox{\bbfvi #1}}\else{\bbfxi #1}\fi}%
\else\ifdim\the\fontdimen6\the\font<13pt   
 {\ifmmode\mathchoice{\mbox{\bbfxii #1}}
  {\mbox{\bbfxii #1}}{\mbox{\bbfviii #1}}
  {\mbox{\bbfvi #1}}\else{\bbfxii #1}\fi}%
\else\ifdim\the\fontdimen6\the\font<15pt   
 {\ifmmode\mathchoice{\mbox{\bbfxiv #1}}
  {\mbox{\bbfxiv #1}}{\mbox{\bbfx #1}}
  {\mbox{\bbfvii #1}}\else{\bbfxiv #1}\fi}%
\else\ifdim\the\fontdimen6\the\font<18pt   
 {\ifmmode\mathchoice{\mbox{\bbfxvii #1}}
  {\mbox{\bbfxvii #1}}{\mbox{\bbfxii #1}}
  {\mbox{\bbfx #1}}\else{\bbfxvii #1}\fi}%
\else\ifdim\the\fontdimen6\the\font<23pt   
 {\ifmmode\mathchoice{\mbox{\bbfxx #1}}
  {\mbox{\bbfxx #1}}{\mbox{\bbfxiv #1}}
  {\mbox{\bbfxii #1}}\else{\bbfxx #1}\fi}%
\else                                      
 {\ifmmode\mathchoice{\mbox{\bbfxxv #1}}
  {\mbox{\bbfxxv #1}}{\mbox{\bbfxx #1}}
  {\mbox{\bbfxvii #1}}\else{\bbfxxv #1}\fi}%
\fi\fi\fi\fi\fi\fi\fi\fi\fi\fi\fi}}

\begin{document}

\title{\Large\bf 
Path-following based Point Matching  using Similarity Transformation} 

\date{}

\author{\begin{tabular}[t]{c@{\extracolsep{4em}}c@{\extracolsep{4em}}c} 
 Wei Lian \\ \\
Dept. of Computer Science,
Changzhi University   \\
Changzhi, Shanxi, P.R. China, 046031  \\
E-mail: lianwei3@foxmail.com 
\end{tabular}}

\maketitle


\thispagestyle{empty}
\subsection*{\centering Abstract}
{\em To address the problem of 3D point matching 
where the poses of two point sets are unknown,
we adapt a recently proposed path following based method to use similarity transformation
instead of the original affine transformation.
The reduced number of transformation parameters leads to 
more constrained and desirable matching results.
Experimental results  demonstrate  better robustness of the proposed method over
 state-of-the-art methods.}

\section{Introduction}

Point matching 
is a  challenging  problem with applications in 
computer vision and pattern recognition. 
To solve this problem,
the robust point matching (RPM) algorithm \cite{RPM_TPS} 
uses deterministic annealing for optimization.
But it needs regularization to avoid  undesirable matching results and
has the tendency of aligning the mass centers of two point sets.
To address this issue,
Lian and Zhang reduced
the objective function of RPM to a  concave function of the point correspondence variable
and  used the branch-and-bound (BnB) algorithm for optimization  \cite{RPM_concave,RPM_model_occlude}.
These methods are  more  robust, 
but their  worse case time complexity is  exponential  due to use of BnB.
To address this issue,
Lian used the path following (PF) strategy \cite{GM_PF_quadratic} to optimize the objective function of \cite{RPM_model_occlude}
by adding a convex quadratic term to the objective function and dynamically changing the weights of the  terms
\cite{RPM_PF_aff}.

But
in the case of 3D matching, 
the method of \cite{RPM_PF_aff} is experimentally shown to only perform well when the transformation is regularized,
while there are problems where the poses of two point sets are unknown 
which call for matching methods where the transformations are not regularized.
The reason 
that the method of \cite{RPM_PF_aff} performs poorly 
is that it uses  affine transformation
which has large number of parameters,
thus resulting in high degree of transformation freedom and unconstrained matching results.
To address this issue, 
we modify  the method to use similarity transformation
whose number of parameters is considerably smaller. 

\section{
The RPM objective function
\label{sec:RPM}
}

Suppose we are to match
two point sets   
\scalebox{1.}{$\mathscr X = \{x_i , 1\le i\le m\}$}
and 
$\mathscr Y = \{y_j , 1\le j\le n\}$ in $\mathbb R^d$.
For this problem, 
RPM uses the following
 mixed linear assignment$-$least square model:
\begin{align}
&\min\ \Phi(P, s,R, t){=}
\sum_{i,j}p_{ij}\| { y}_j-sR  x_i {-}  t\|^2 
-\mu 1_m^\top P1_n \notag\\
 &\quad= 1_m^\top P  {\widetilde y} + s^2 {\widetilde x}^\top P  1_n 
 - 2s\cdot \text{tr}(R X^\top P Y )
 + 1_m^\top P1_n \|  t\|^2  
   \notag \\
&\qquad -2  t^\top (Y^\top P^\top  1_m   - s R X^\top P   1_n)  
-\mu 1_m^\top P1_n  \label{E_lin_tr} \\
 &s.t.\ \  P 1_n\le1_m, \quad 1_m^\top P\le 1_n,   
 \quad P\ge 0, \label{inequal_assign_const} \\
&\ \qquad \underline{s}\le s \le \overline{s} \label{s_range}
\end{align}
Here we use
similarity transformation
with $R$, $s$ and $t$ being rotation matrix,
scale change and translation vector. 
The constants $\underline{s}\ge 0$ and $\overline{s}\ge0$ are lower and upper  bounds of $s$.
The matching matrix $P=\{p_{ij}\}$ has   $p_{ij}=1$
if two points $i$, $j$ are matched and $0$  otherwise.
The last term in $\Phi$ is used to regularize the number of correct matches 
with $\mu$ being the balancing weight.
$\|\cdot\|$ is  the $l_2$  norm of a vector
and tr() denotes the trace of a square matrix.
$1_n$ represents the $n$-dimensional vector of all ones.
The matrices
$X\triangleq\begin{bmatrix}
    x_{1}, \dots,  x_{m}
  \end{bmatrix}^\top$,
$Y\triangleq\begin{bmatrix}
    y_{1}, \dots,  y_{n}
  \end{bmatrix}^\top$
and vectors    $ {\widetilde x}\triangleq \begin{bmatrix}
   \| x_1\|^2, \dots, \| x_{m}\|^2
  \end{bmatrix}^\top$,
$ {\widetilde y}\triangleq \begin{bmatrix}
   \| y_1\|^2, \dots, \| y_{n}\|^2
  \end{bmatrix}^\top$.

It's easily seen  that given the values of $P$, $s$ and $R$,
$\Phi$ is a convex quadratic function of $t$.
Hence,
the optimal   $t$  minimizing $\Phi$ can be obtained via  $\frac{\partial \Phi}{\partial { t}}=0$ to be
\scalebox{1.1}{
$
\widehat { t}
 =
  \frac{1}{1_m^\top P1_n }( Y^\top P^\top  1_m  -  s R X^\top P  1_n)
$}.
Substituting  $\widehat { t}$  into $\Phi$ to eliminate $ t$ yields
an energy function with reduced number of variables: 
\begin{align}
&\Phi(P,s,R)=
 1_m^\top P  {\widetilde y} -\mu 1_m^\top P 1_n - \frac{1}{1_m^\top P1_n} \|Y^\top P^\top  1_m\|^2  \notag\\
&+ s^2 ( {\widetilde x}^\top P  1_n   - \frac{1}{1_m^\top P1_n}    \| X^\top P  1_n\|^2 )  \notag\\
&- 2s\cdot\text{tr}(R ( X^\top P Y - \frac{1}{1_m^\top P1_n} X^\top P 1_n 1_m^\top PY ))  \label{E_P_s_R}
\end{align}

\section{Optimal $s,R$ minimizing $\Phi(P,s,R)$ \label{optimal_s_R}}
Let matrix
\[
 A\triangleq   
  X^\top P Y - \frac{1}{1_m^\top P1_n} X^\top P 1_n 1_m^\top PY 
\]
and let $USV^\top$ be the singular value decomposition of $A^\top$,
where $S$ is a diagonal matrix and
the columns of  $U$ and $V$ are orthogonal unity vectors.
Then given $s> 0$, 
the optimal rotation matrix  $R$ minimizing $\Phi$ in   \eqref{E_P_s_R} is
$\widehat R=U\text{diag}(\begin{bmatrix}
                     1,\ldots,1,\det(UV^\top)
                    \end{bmatrix})V^\top$ \cite{CPD},
where $\text{diag}(\cdot)$ denotes converting a vector into a diagonal matrix
and $\det(\cdot)$ is the determinant of a square matrix.
Substituting $\widehat R$  into \eqref{E_P_s_R} to eliminate $R$ 
yields  a (possibly concave) quadratic program in single variable $s$.
Given the range of $s$ as $\underline{s}\le s \le \overline{s}$,
one can easily solve  this  quadratic program 
by comparing the function values at the boundary points $\underline{s}$, $\overline{s}$ and 
the extreme  point.

\section{An objective function in one variable $P$
\label{sec:regularize}}

We aim to obtain an objective function only in one variable $P$,
which can be achieved by minimizing $\Phi$ with respect to $s$ and $R$, i.e.:
\begin{align}
& \Phi(P)\triangleq\min_{s,R}  \Phi(P,s,R) 
\label{E_P}
\end{align}

  
For $\Phi(P)$, 
the following results can be established:

\begin{proposition} \label{Prop_one}
$\Phi(P)$ is  concave under constraints \eqref{s_range}. 
\end{proposition}

\begin{figure}[h]
\centering
 \includegraphics[width=0.5\linewidth]{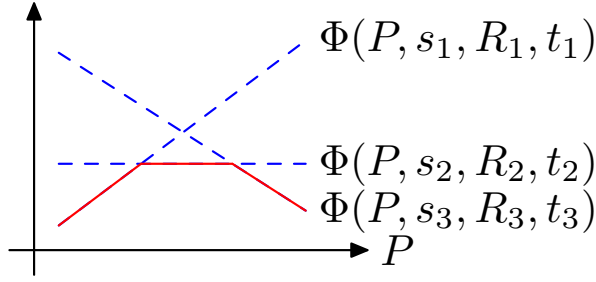}  
 \caption{ 
Point-wise minimization of a family of linear functions $\Phi(P,s,R,t)$ (dashed  straight lines)
with respect to parameters $s$, $R$ and $t$ results in a concave function (solid piecewise straight line).
\label{linear_min}
}
\end{figure}

{\proof
Based on the aforementioned derivation, 
we have $\Phi(P,s,R )=\min_{{ t}} \Phi (P,s,R,{ t})$.
Consequently, we have  
\[\Phi(P)=\min_{s,R} \Phi(P,s,R)=
\min_{s,R,{ t}} \Phi(P,s,R,{ t})\]
For each $s$, $R$ and ${ t}$, 
$\Phi(P,s,R,{ t})$ is apparently  a linear function of $P$.
We see that $\Phi(P)$ is the point-wise minimum of a family
of linear functions,
and thus is concave, as illustrated in Fig. \ref{linear_min}.
}

The fact that $\Phi(P)$ is concave makes it easier for 
the PF algorithm to be applied to the minimization of our objective function
as it requires two terms, a concave and a convex term, to be provided.

\begin{proposition}
There exists an integer solution for any local minima (including the global minimum) of function
$\Phi(P)$ under constraints \eqref{inequal_assign_const} and \eqref{s_range}.
\end{proposition}

{\proof
The polytope formed by  constraint \eqref{inequal_assign_const}
satisfies the total unimodularity property \cite{book_comb_optimize},
which means that the coordinates of the vertices of this polytope are integer valued.
We already proved that $\Phi(P)$ is concave under constraints 
\eqref{s_range}. 
It is well known that any local minima 
(including the global minimum)
of a concave function over a polytope can be obtained at one of its vertices.
Thus, the proposition follows.

}

This result implies that minimization of $\Phi(P)$ 
by simplex-like algorithms results in integer valued solution.
This is important  as it avoids the need of discretizing solutions
which can cause error
and poor performance 
\cite{LP_edge}. 


To facilitate optimization of $\Phi$, 
we needs to convert  $P$  into a vector.
We define the vectorization of a matrix as the concatenation of its rows
\footnote{This is different from the conventional definition.},
denoted by $\text{vec}()$.
Let $p\triangleq\text{vec}(P)$.
To get the form of $\Phi$ in terms of vector $p$,
new denotations are needed.
Let
\begin{gather}
 \text{vec}(X^\top PY)\triangleq B  p,\
 X^\top P 1_n\triangleq C  p, \
 Y^\top P^\top  1_m\triangleq D p,    \notag\\ 
  {\widetilde x}^\top P  1_n\triangleq   a^\top  p, \quad
 1_m^\top P \widetilde Y \triangleq b^\top p  \notag
\end{gather}
Based on the fact $\text{vec}(M_1 M_2 M_3)= (M_1\otimes M_3^\top)\text{vec}(M_2)$ for any matrices  $M_1$, $M_2$ and $M_3$,
we have matrices
$
B=X^\top\otimes Y^\top,\ C=X^\top\otimes  1_n^\top,\ D= 1_m^\top\otimes Y^\top  \notag
$
 and vectors
\scalebox{1.1}{
$
  a= {\widetilde x} \otimes  1_n,\
b=1_m \otimes \widetilde y  \notag
$}.
Here $\otimes$ denotes the Kronecker product.
With the above preparation, 
$\Phi(P)$  can be written in terms of vector $ p$ as
\begin{align}
\Phi( p)=&  
(b-\mu 1_{mn})^\top   p  - \frac{1}{1_{mn}^\top p} \|D  p \|^2  
 + \min_{s,R} \{s^2 ( a^\top  p    - \frac{1}{1_{mn}^\top p}    \| C  p \|^2 )  \notag\\
&- 2s\cdot \text{tr}(R [ \text{mat} (B p)  - \frac{1}{1_{mn}^\top p} C p  p^\top D^\top ]) \} \label{phi_p}
\end{align}
where $\text{mat}()$ denotes converting a vector into a matrix,
which can be seen as  inverse of the operator  $\text{vec}()$.

To facilitate  optimization of $\Phi$, 
we need to get  the formula of the gradient of $\Phi$. 
As  $\Phi$ involves minimization operations, 
it's difficult to directly derive the formula of \scalebox{1.}{$\frac{\partial \Phi}{\partial p}$}.
To address this issue,
we appeal to the result of Danskin's theorem \cite{cov_analysis} (page 245 therein), 
which in our case states that if $\Phi(p,s,R)$ is concave in $p$ for each $s$ and $R$
(this can be proved  analogously as the proof of Proposition \ref{Prop_one})
and the feasible regions of $s$ and $R$ are compact,
then 
\scalebox{1.1}{
$\Phi(p)= \min_{s,R} \Phi(p, s, R)$}
has gradient:
\begin{align}
&\frac{\partial\Phi(p)}{\partial p}= \frac{\partial\Phi(p,\widehat s,\widehat r)}{\partial p}=         
b -\mu 1_{mn} 
-\frac{2}{1_{mn}^\top p} D^\top D p    \notag\\
&+\frac{\|Dp\|^2 }{(1_{mn}^\top p)^2}1_{mn} 
+  \widehat s^2 ( a    - \frac{2}{1_{mn}^\top p}   C^\top C  p  + \frac{1}{(1_{mn}^\top p)^2}    \| C  p \|^2 1_{mn})  \notag\\
&- 2\widehat s \{ B^\top \widehat r  -  \frac{1}{1_{mn}^\top p}  [D^\top (p^\top C^\top  \otimes I_d) + C^\top(I_d\otimes p^\top D^\top) ] \widehat r \notag\\
&+ \frac{1}{(1_{mn}^\top p)^2} \text{tr}( \widehat R  C p  p^\top D^\top ) 1_{mn} 
\}  \label{gradient_exp}
\end{align}
where  $\widehat s$ and $\widehat R$ 
satisfy
\scalebox{1.1}{
$
\Phi(p,\widehat s,\widehat R)= \min_{s,R} \Phi(p, s, R)
$}.
The optimal $\widehat s$ and $\widehat R$ can be obtained by the method described previously. 
Here the vector $\widehat r\triangleq\text{vec}(\widehat R^\top)$ and
$I_{d}$ denotes the $d\times d$ identity matrix.

\section{PF based optimization\label{sec:optimize}}
The PF algorithm \cite{GM_PF_quadratic} is used to optimize $\Phi$
 by constructing an interpolation function 
between a convex  function $\|p\|^2$ and the concave function $\Phi$,
\[
 E_\lambda=(1-\lambda)\|p\|^2 +\lambda \Phi(p)
\]
and gradually increasing $\lambda$ from $0$ to $1$ so that
$E_\lambda$ gradually transitions from the convex function $\|p\|^2$ to the concave function $\Phi$.
With each value of $\lambda$, $E_\lambda$ is locally minimized.
We refer the reader to \cite{RPM_PF_aff} for detail.

\section{Experimental results}
We compare our method with state-of-the-art methods including
RPM-PF~\cite{RPM_PF_aff}, RPM~\cite{RPM_TPS}, 
Go-ICP~\cite{Go_ICP}, CPD~\cite{CPD} and gmmreg~\cite{kernel_Gaussian_journal}.
To ensure fairness, for RPM-PF,
transformation is not regularized.
We implement all the  methods in MATLAB on a
PC with a 3.3 GHz CPU and 16 G RAM. For methods only
outputting point correspondence, affine transformation is used
to warp the model point set. For our method, we set parameters
$\underline{s}=0.5$ and $\overline{s}=1.5$.

\begin{figure}[t]
 \includegraphics[width=\linewidth]{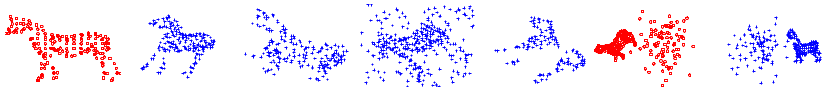}

 
\caption{
First 5 columns:
model point set (left column) and 
examples of scene  sets in the deformation, positional noise,  outlier and  occlusion tests, respectively (columns 2 to 5).
Last 2 columns:
examples of model (column 6) and scene (right column) point sets in the coexisting outlier test.
\label{3D_test_exa}}
\end{figure}

Following \cite{GM_relax_label,RPM_model_occlude},
we test a method's robustness to non-rigid deformation, positional noise,
outliers, occlusion and coexisting outliers,
as illustrated in the second to fifth column of Fig. \ref{3D_test_exa}.
Also, to test a method's ability to handle rotation and scale changes,
random rotation with rotation angle less than 60 degree 
and random uniform scaling with scale factor within range $[0.5 ,1.5]$
are applied to the prototype shape 
when generating the scene point set.


\begin{figure}[!ht]

\centering
{\includegraphics[width=0.325\linewidth]{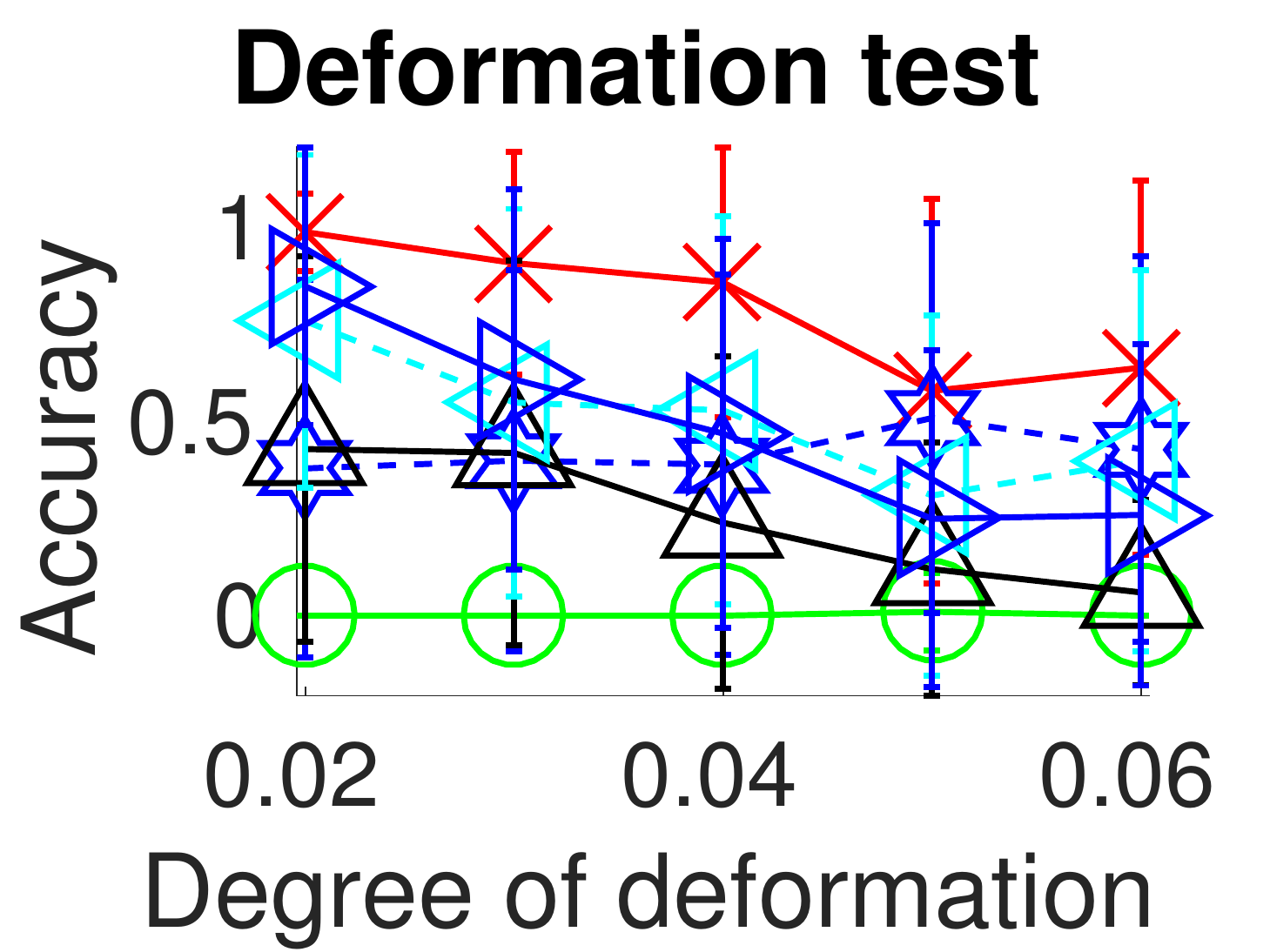}}
{\includegraphics[width=0.325\linewidth]{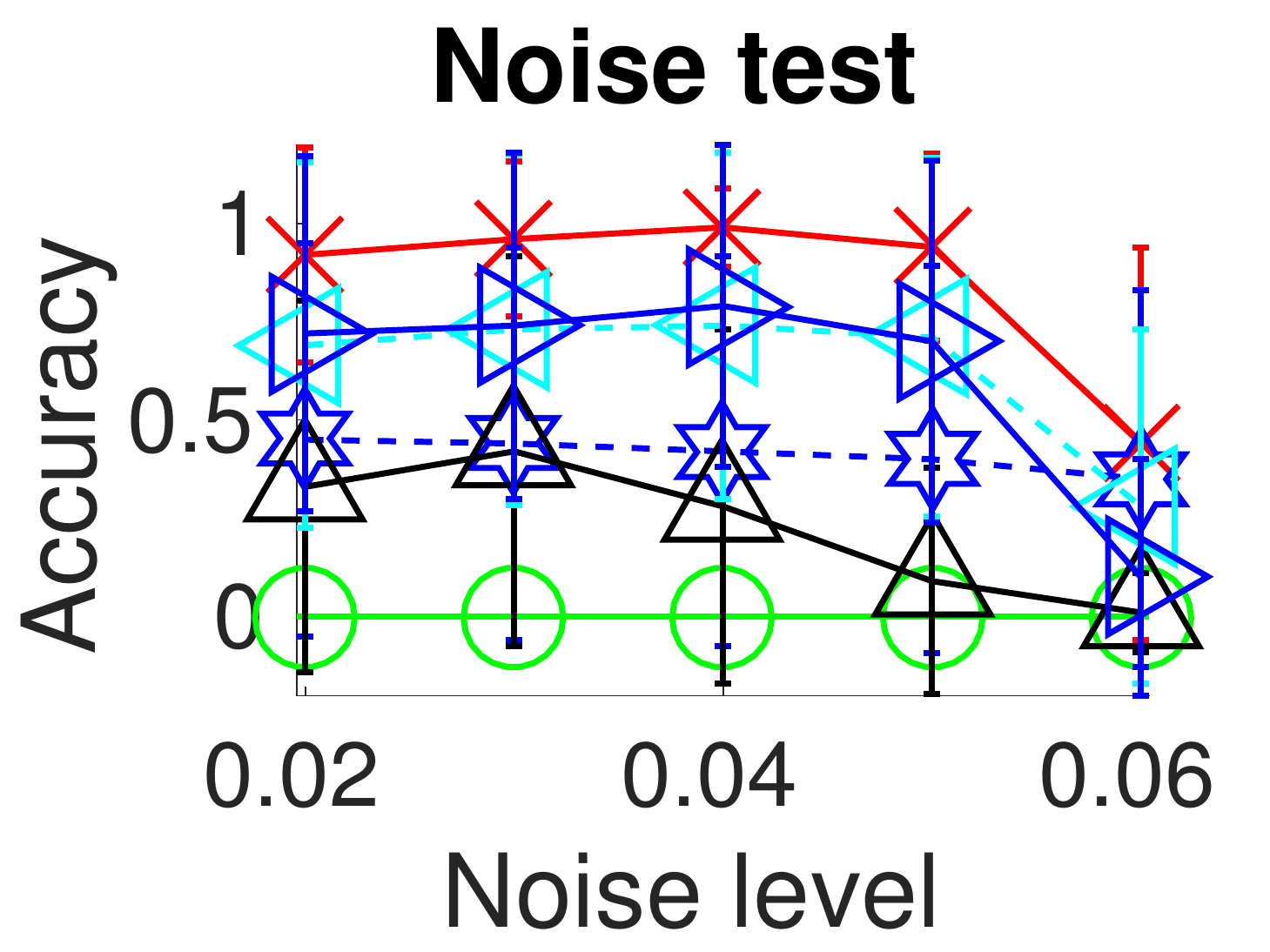}}
{\includegraphics[width=0.325\linewidth]{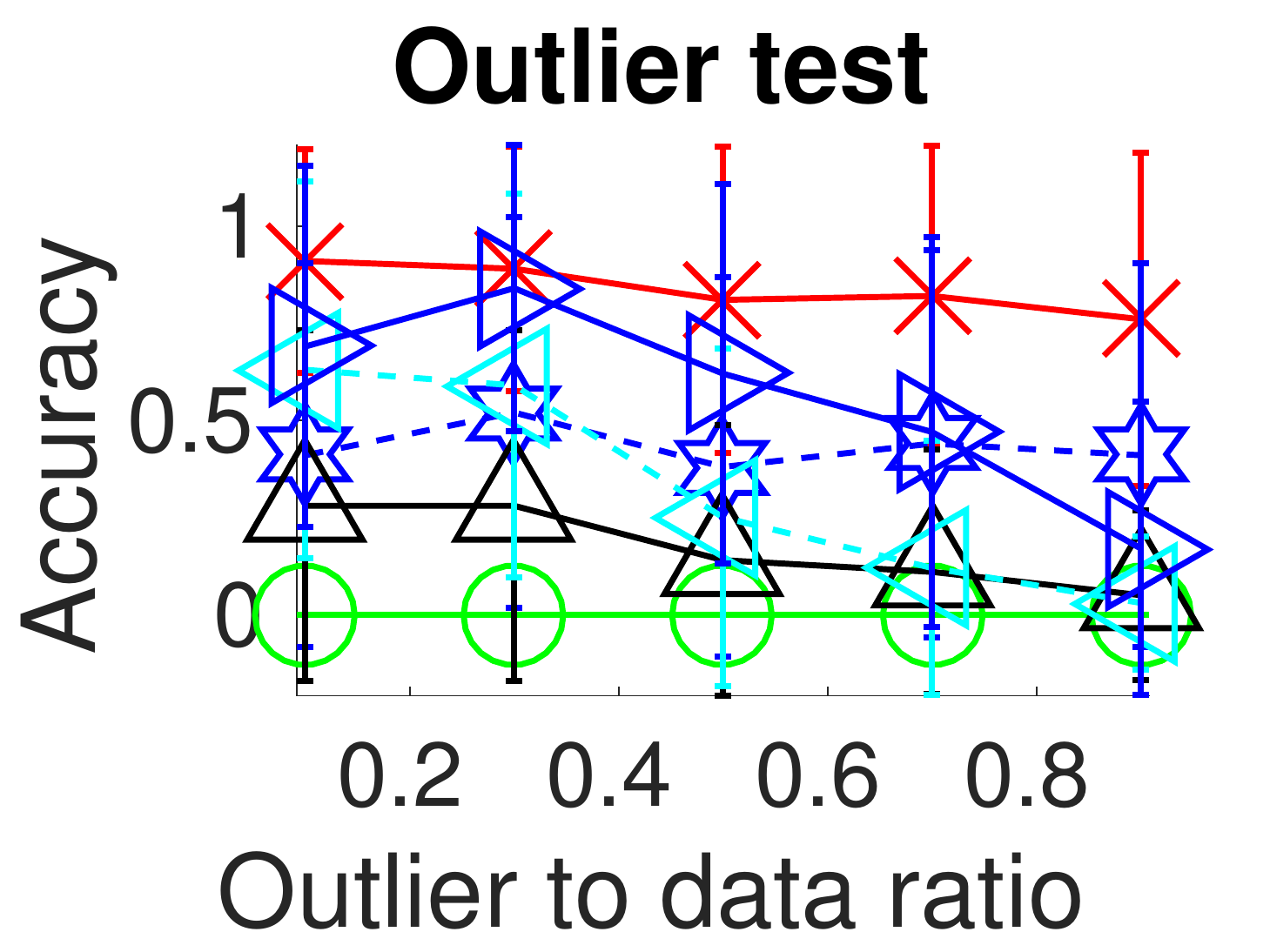}}

{\includegraphics[width=0.33\linewidth]{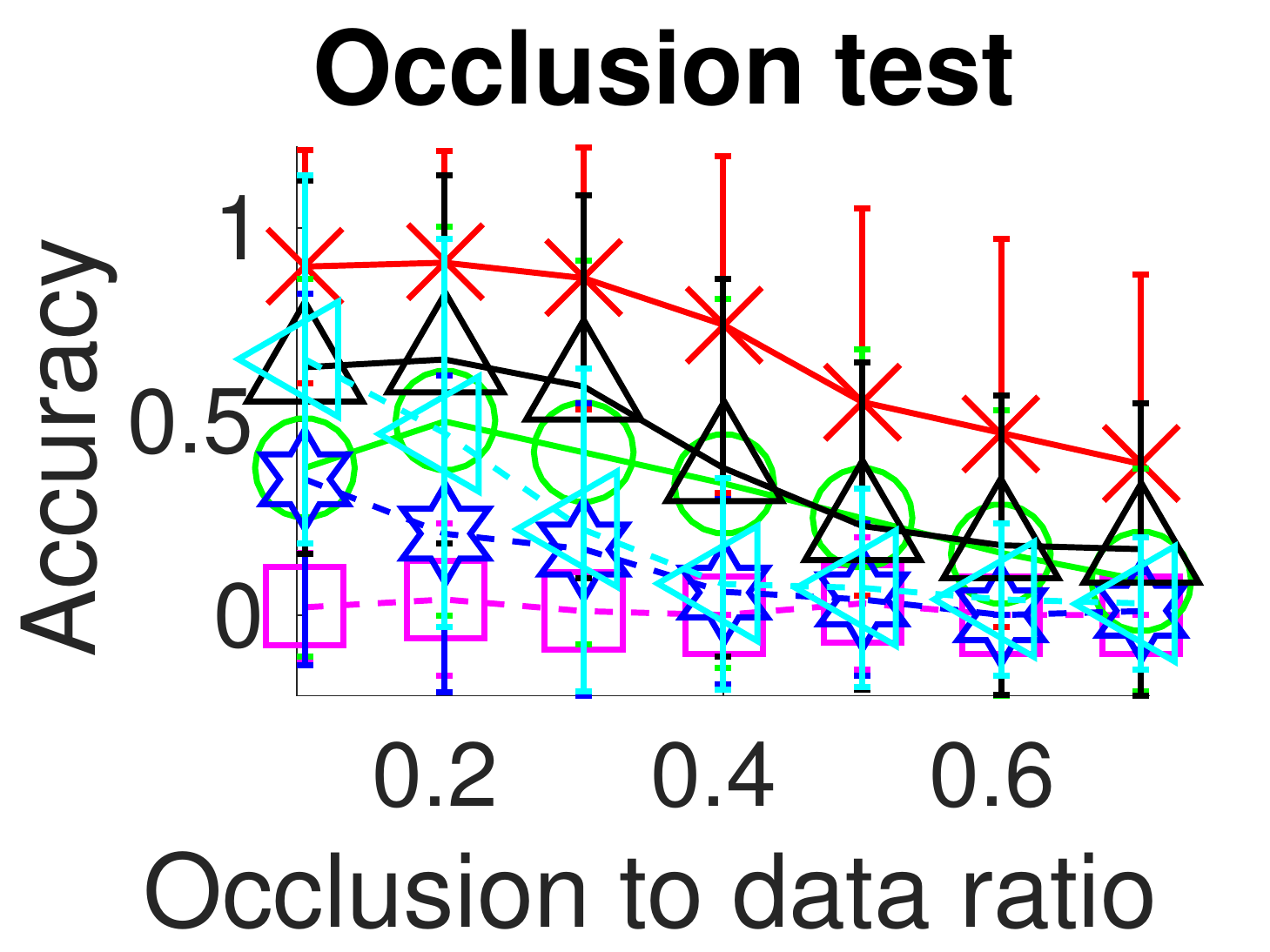}}
{\includegraphics[width=0.46\linewidth]{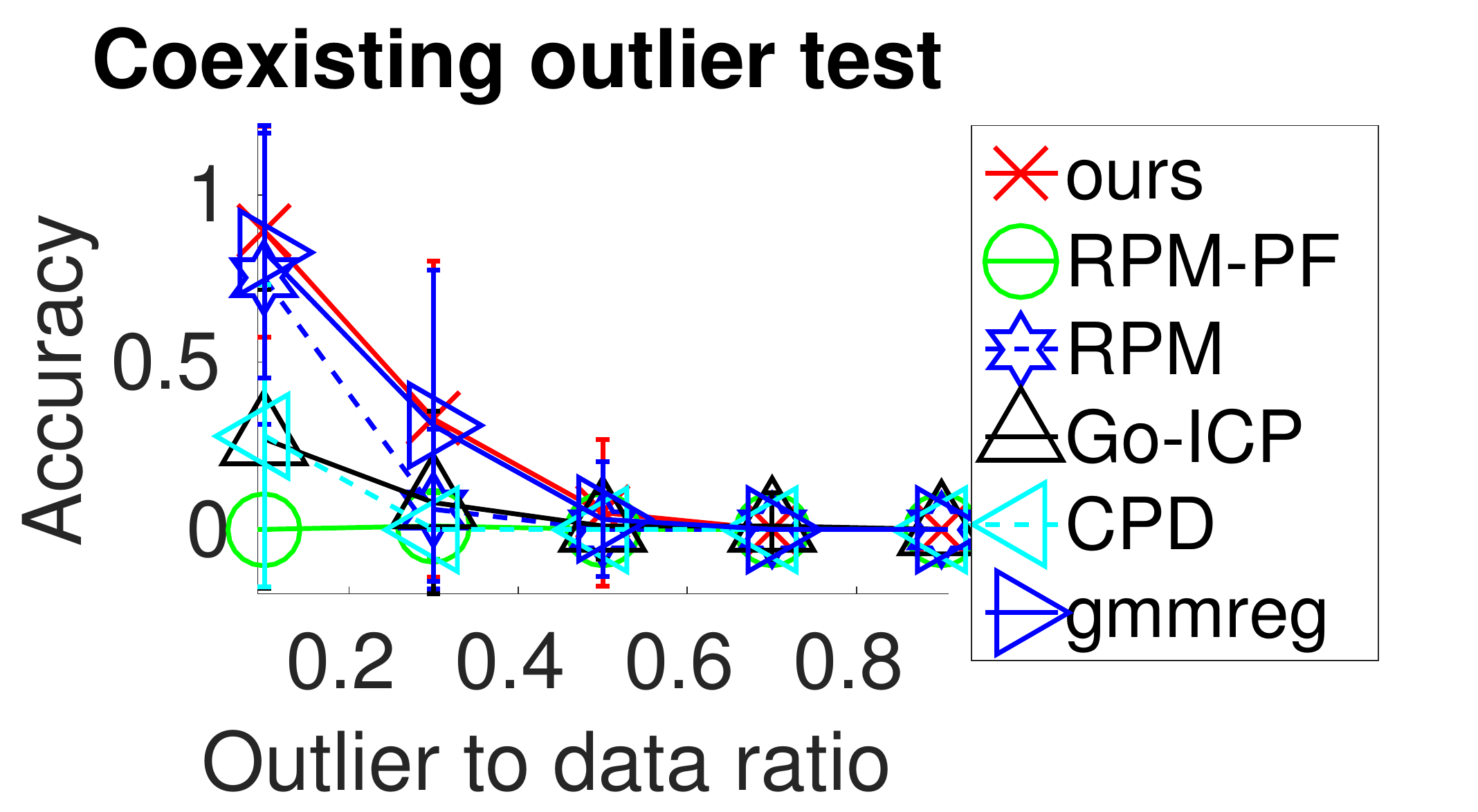}}
\caption{
Average matching accuracies  by different methods in the 5 categories of tests.
The error bars indicate standard deviations of the methods over 100 random trials.
\label{3D_test_statis}
}
\end{figure}

\begin{figure}[!ht]
\centering

\includegraphics[width=\linewidth]{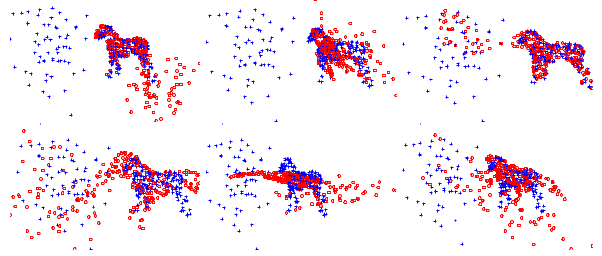}

\caption{
Examples of matching results by (from left to right and from top to bottom)
our method, RPM-PF, RPM, Go-ICP, CPD and gmmreg
 in the  coexisting outlier  test.
\label{3D_match_exa}
}
\end{figure}

The average matching accuracies (fraction of correct matches)
by
different 
methods  are presented in Fig. \ref{3D_test_statis}. 
One can see that our method performs considerably better than  other methods.
This demonstrates our method's robustness to various types of disturbances.
Examples of matching results by different methods in the 
coexisting outlier test are shown  in Fig. \ref{3D_match_exa}.
The average running times (in second) by different methods are 
   3.6054 for our method,
    4.1525 for RPM-PF,
    1.8160 for RPM,
    5.7997 for Go-ICP,
    0.0612 for CPD and
    0.2865 for gmmreg.
It's clear our method is  efficient.


    
%

\section{Conclusion}
We proposed a PF based point matching method in this letter
by adapting the method of \cite{RPM_PF_aff} to use the similarity transformation.
Due to nonlinearity of 3D similarity transformation,
this is a nontrivial extension of the method of \cite{RPM_PF_aff}.
Experimental results demonstrate better robustness of the proposed method 
over  state-of-the-art methods.

%



\begin{thebibliography}{}
\bibitem{RPM_TPS}
Chui, H., Rangarajan, A.:
`A new point matching algorithm for non-rigid registration'.
\textit{Computer Vision and Image Understanding},
2003,  \textbf{89}, pp. 114-141

\bibitem{RPM_concave}
Lian, W., Zhang, L.:
`Robust point matching revisited: a concave optimization approach'. 
\textit{European conference on computer vision}, 2012


\bibitem{RPM_model_occlude} 
Lian, W., Zhang, L.:
`Point matching in the presence of outliers in both point sets: A concave optimization approach',
\textit{IEEE Conf. Computer Vision and Pattern Recognition}, 2014, pp.  352-359


\bibitem{GM_PF_quadratic}
Liu, Z. Y. and Qiao, H.:
`Gnccp-graduated nonconvexity and concavity procedure'. 
\textit{IEEE Trans. Pattern Analysis and Machine Intelligence},
2014,  \textbf{36}, pp. 1258-1267

\bibitem{RPM_PF_aff}
Lian, W.: 
`A path-following algorithm for robust point matching'. 
\textit{IEEE Signal Processing Letters}, 
2015, \textbf{23}, pp. 89-93

\bibitem{CPD}
Myronenko, A., Song, X.:
`Point set registration: Coherent point drift'.
\textit{IEEE Transactions on Pattern Analysis and Machine Intelligence},
2010, \textbf{32}, pp. 2262-2275

\bibitem{book_comb_optimize} Papadimitriou, C.H., Steiglitz, K.:
`Combinatorial optimization: algorithms and complexity',
\textit{Dover Publications, INC. Mineola. New York}, 
1998


\bibitem{LP_edge}
Jiang, H.,  Drew, M. S., Li, Z. N.:
`Matching by linear programming and successive convexification'.
\textit{IEEE Trans. Pattern Analysis and Machine Intelligence},
2007, \textbf{29}, pp. 959-975


\bibitem{cov_analysis}
Bertsekas, D. P.:
`convex analysis and optimization'.
Athena Scientific,
Belmont, Massachusetts, 2003

\bibitem{Go_ICP}
Yang, J., Li, H., Jia, Y.:
`Go-icp: Solving 3d registration efficiently and globally optimally'.
\textit{IEEE International Conference on Computer Vision}, 
2013

\bibitem{kernel_Gaussian_journal}
Jian, B., Vemuri, B. C.:
`Robust point set registration using gaussian mixture models'.
\textit{IEEE Trans. Pattern Analysis and Machine Intelligence},
2011, \textbf{33}, pp. 1633-1645

\bibitem{GM_relax_label}
Zheng, Y., Doermann, D.:
`Robust point matching for nonrigid shapes by preserving local neighborhood structures'.
\textit{IEEE Trans. Pattern Analysis and Machine Intelligence},
2006, \textbf{28}, pp. 643-649

%
%



%















\end{thebibliography}

\end{document}